\documentclass[lettersize,journal]{IEEEtran}
\usepackage{amsmath,amsfonts}
\usepackage{algorithmic}
\usepackage{algorithm}
\usepackage{array}
\usepackage[caption=false,font=normalsize,labelfont=sf,textfont=sf]{subfig}
\usepackage{textcomp}
\usepackage{stfloats}
\usepackage{url}
\usepackage{verbatim}
\usepackage{graphicx}
\usepackage{cite}
\usepackage{graphicx}
\usepackage{amsmath}
\usepackage{amsthm}
\usepackage{booktabs}
\usepackage{algorithm}
\usepackage{algorithmic}
\usepackage{bm}
\usepackage{diagbox}
\usepackage{makecell}
\usepackage{CJKutf8}
\usepackage{overpic}
\usepackage{tabularx}

\begin{document}

\title{Hand Hygiene Assessment via Joint Step Segmentation and Key Action Scorer}

\author{Chenglong Li, Qiwen Zhu, Tubiao Liu, Jin Tang, and Yu Su
\thanks{This work is partly supported by National Natural Science Foundation of China (No. 62076003), and Anhui Provincial Key Research and Development Program (No. 202104d07020008).

C. Li is with Information Materials and Intelligent Sensing Laboratory of Anhui Province, Anhui Provincial Key Laboratory of Multimodal Cognitive Computation, School of Artificial Intelligence, Anhui University, Hefei 230601, China. (Email: lcl1314@foxmail.com)

Q. Zhu, T. Liu and J. Tang are with Information Materials and Intelligent Sensing Laboratory of Anhui Province, Key Lab of Intelligent Computing and Signal Processing of Ministry of Education, Anhui Provincial Key Laboratory of Multimodal Cognitive Computation, School of Computer Science and Technology, Anhui University, Hefei 230601, China. (Email: zqw\_0327@126.com, 1987055525@qq.com, tangjin@ahu.edu.cn)

Y. Su is with Hefei Normal University and Institute of Artificial Intelligence, Hefei Comprehensive National Science Center, Hefei 230061, China. (Email: yusu@hfnu.edu.cn)
}
}

\markboth{Journal of \LaTeX\ Class Files,~Vol.~14, No.~8, August~2021}%
{Shell \MakeLowercase{\textit{et al.}}: A Sample Article Using IEEEtran.cls for IEEE Journals}


\maketitle

\begin{abstract}
Hand hygiene is a standard six-step hand-washing action proposed by the World Health Organization (WHO). However, there is no good way to supervise medical staff to do hand hygiene, which brings the potential risk of disease spread. 
Existing action assessment works usually make an overall quality prediction on an entire video. However, the internal structures of hand hygiene action are important in hand hygiene assessment.
Therefore, we propose a novel fine-grained learning framework to perform step segmentation and key action scorer in a joint manner for accurate hand hygiene assessment.
Existing temporal segmentation methods usually employ multi-stage convolutional network to improve the segmentation robustness, but easily lead to over-segmentation due to the lack of the long-range dependence. To address this issue, we design a multi-stage convolution-transformer network for step segmentation.
Based on the observation that each hand-washing step involves several key actions which determine the hand-washing quality, we design a set of key action scorers to evaluate the quality of key actions in each step. 
In addition, there lacks a unified dataset in hand hygiene assessment. Therefore, under the supervision of medical staff, we contribute a video dataset that contains 300 video sequences with fine-grained annotations. 
Extensive experiments on the dataset suggest that our method well assesses hand hygiene videos and achieves outstanding performance. 
\end{abstract}

\begin{IEEEkeywords}
Hand Hygiene, Action Assessment, Step Segmentation, Key Action, Joint Learning.
\end{IEEEkeywords}

\section{Introduction}
\IEEEPARstart{I}{n} 2005, the World Health Organization designated October 15-th as "World Hand-washing Day". But not many people can wash their hands well in life, and quite a few people have not developed good hand-washing habits. Taking novel coronavirus as an example, since the outbreak, more than 500 million people have been diagnosed and more than 6 million people have died in the world. The droplets sprayed by patients with the virus will not only spread into the air but also stick to their hands and survive for a long time. 
If they touch other persons or other things with their hands again, they will have a great chance of infecting other persons. Therefore, the hand is also an important medium for these viruses to spread diseases. Research shows that scientific hand washing can reduce the risk of illness by 20\%. Therefore, it is very important to have a correct assessment for the hand-washing process. With the assessment, medical and other staff can correct their hand-washing actions to reduce the risk of disease spread as much as possible. 

\begin{figure}[t]
    \centering
    \includegraphics[width=1\linewidth,height=0.9\columnwidth]{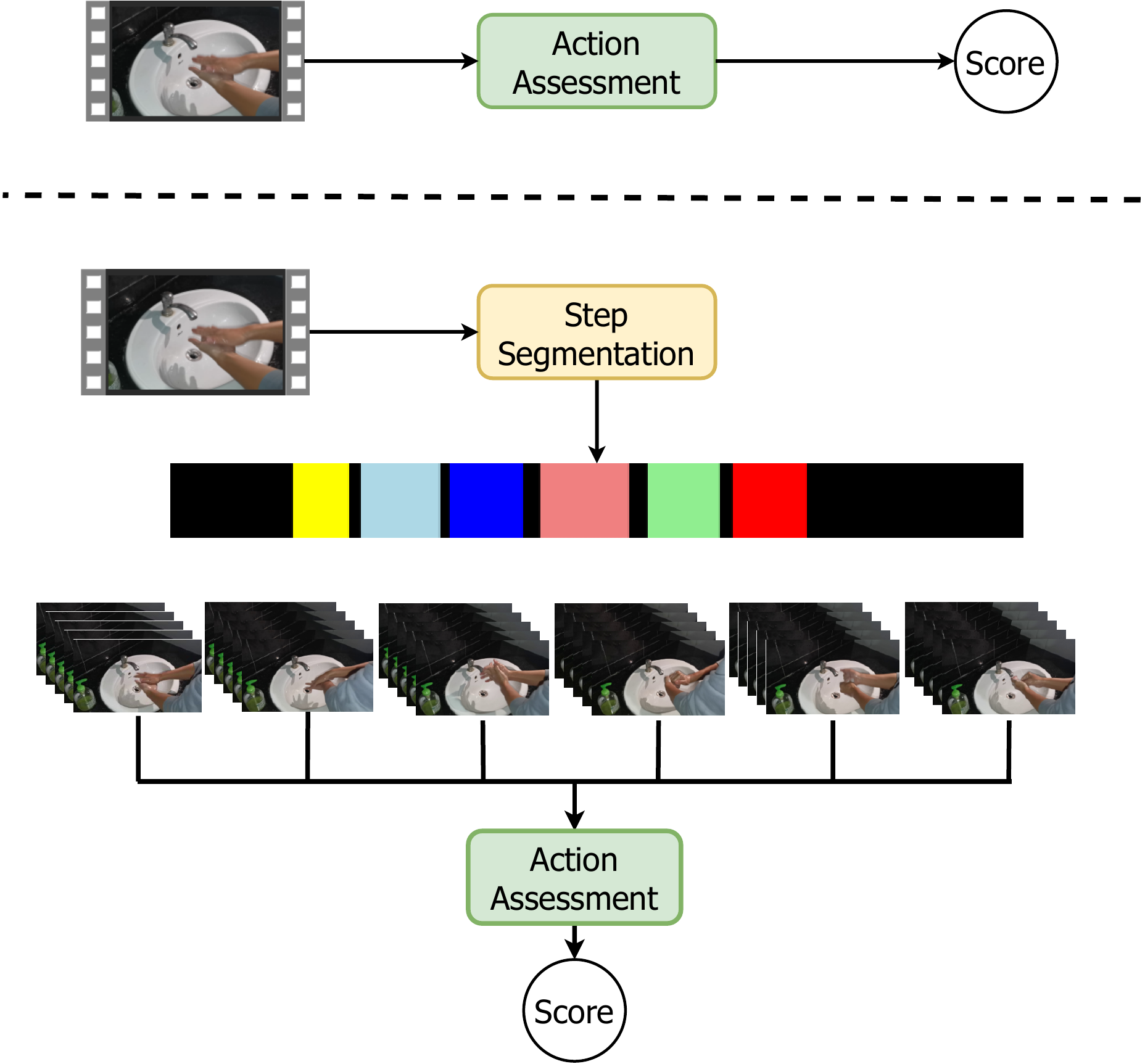}
    \caption{Comparison of traditional action assessment framework with our framework for hand hygiene assessment. The existing framework~\cite{Zeng,Yu} directly predicts a score for a video, while we first use the step segmentation model to segment each step in the video, and then accurately assess them one by one by evaluating their key actions.}
    \label{fig:contrast}
\end{figure}

In recent years, action assessment, which aims to evaluate the quality of action, has attracted extensive attention~\cite{Pirsiavash,Parmar,Li1,xu2019learning,zhang2022semi,xu2022finediving,jain2020action}. It has important applications in many fields in the real world, such as sports and medical treatment. Previous action assessment models~\cite{Zeng,Yu} often directly evaluate a video and compute a score, as the top in Fig.~\ref{fig:contrast}. However, these methods would ignore many details in long actions which often involve several short steps, and thus has degraded performance in long action assessment. 

The hand hygiene stipulated by the World Health Organization (WHO) is a standard long action, which includes six steps. There are three major issues in the task of hand hygiene assessment. 
First, each video contains some or all of the six steps, and thus existing action assessment models could not know which frames belong to which step, which brings a big challenge to accurately assess the whole video. Second, in each step, there are several key actions to determine the quality, and thus not all actions are useful for hand hygiene assessment. Finally, there is currently a lack of a unified and high-quality annotated video dataset, which limits the research and development in hand hygiene assessment. Since there are many concepts in hand hygiene assessment, we summarize them in Table~\ref{tab:concept}.

\begin{table}[h]
  \centering
  \caption{Concept explanations in hand hygiene assessment.}
  \begin{tabularx}{\columnwidth}{lX}
    \toprule
     \textbf{Concept}             & \textbf{Explanation}   \\
    \midrule
    {\textbf{Long action}} &The whole action includes several steps  and the total \newline time is more than 10 seconds, such as hand hygiene. \\
    {\textbf{Short action}} &The whole action includes fewer steps and the total \newline duration is less than 10 seconds, such as diving. \\
    {\textbf{Step}}   & The certain stage in the hand hygiene assessment task. \\
    {\textbf{Segment}}   &A set of consecutive frames belonging to a step. \\
    {\textbf{Key action}}   &The action that determines the quality of the segment in \newline each step, such as finger crossing and hand changing in \newline the second step.  \\
    \bottomrule
  \end{tabularx}
  \label{tab:concept}
\end{table}


To solve the first two issues, we propose a fine-grained learning framework based on a novel approach of joint step segmentation and key action scorer for robust hand hygiene assessment. 
Most of existing methods~\cite{Zeng,Yu} evaluate short action such as diving by an overall prediction, which can not perform accurate evaluation of long actions since their quality depends on the completion quality of each step. Different from the overall assessment of a whole video in existing action assessment methods, we partition the input video into the step-based video segments for fine-grained scoring. Xu et al.~\cite{xu2022finediving} propose to segment a set of procedures for evaluation, and each two consecutive procedures does not contain the interval, i.e., the frames not belonging to any procedure. However, each two consecutive steps in hand hygiene videos usually has the interval, and the procedure segmentation is unsuitable for our task.
To this end, we design a multi-stage convolution-transformer network for accurate step segmentation in hand hygiene videos. The actions in hand hygiene videos are continuous and similar, and existing multi-stage models~\cite{Yazan,Li} easily lead to over-segmentation due to the lack of the long-range dependence.
To handle this problem, we embed the linear transformer in the multi-stage convolution network to form the multi-stage convolution-transformer network, which establishes effective long-range dependence between frames without increasing too much computation.

\begin{figure}[t]
    \centering
    \includegraphics[width=1\linewidth,height=0.5\columnwidth]{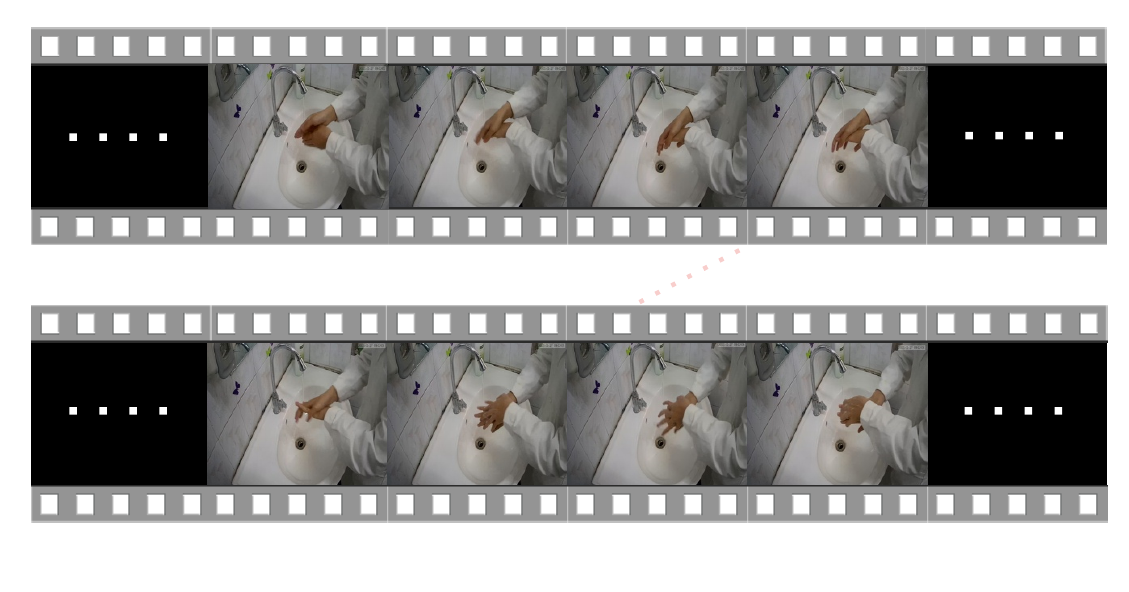}
    \caption{Illustration of the key actions in step 2 from a hand hygiene video.}
    \begin{picture}(0,0)
        \put(-65,95){\scriptsize{(a) First key action: Palm over dorsum}}
        
        \put(-65,35){\scriptsize{(b) Second key action: Fingers interlaced}}

    \end{picture}
    \label{fig:keyaction}
\end{figure}
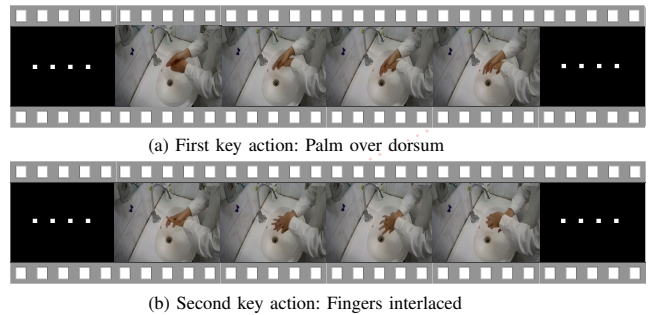

Accurately assessing each step is the most critical issue in hand hygiene assessment. We observe that each step involves two key actions which dominate the quality of this step, as shown in Fig.~\ref{fig:keyaction} and Table~\ref{tab:keyactionofstep}.
Therefore, we propose a new assessment module including six key action scorers, each of which corresponds to a step and is composed of two branches with the same structure. In particular, each branch includes a fully connected layer (FC) and a learnable Sigmoid, and different branches have independent parameters to model the characteristics of different key actions. 

\begin{table}[]
    \centering
    \caption{Key actions in all steps.}\label{tab:keyactionofstep}
    \begin{tabular}{lcc}
        \hline
         Step   &Key action 1 &Key action 2\\
        \hline
         Step 1 & Palms facing each other & Finger together\\
         Step 2 & Palm over dorsum   &Fingers interlaced\\
         Step 3 & Palm to palm   &Fingers interlaced  \\
         Step 4 & Flex fingers  & Rub in the palm\\
         Step 5 & Hold your left thumb &Rotary rub  \\
         Step 6 & Finger together &Rub fingertips in palms  \\
         \hline
    \end{tabular}
\end{table}

To provide a unified platform for hand hygiene assessment, we create a unified and high-quality video dataset, called HHA300. HHA300 contains 300 hand-washing videos of different persons from a wide range of viewpoints and background complexities,  including all possible situations in real-world hand hygiene. To complete the task of step segmentation and action assessment at the same time, we provide two different forms of annotations for HHA300. In specific, We formulate a series of rules to ensure the fairness and quality of annotations, and score each hand hygiene video according to the rules. Besides, to provide high-quality fine-grained annotations, in addition to a total assessment score for each video, we also annotate the frame-level labels, including the six step labels stipulated by WHO and background label, under the supervision of medical staff.

We evaluate our framework on HHA300 dataset including step segmentation and action assessment. The results demonstrate that the joint step segmentation and key action scorer yields a notable performance gain against with other methods. In summary, our contributions are as follows.

\begin{itemize}
    \item  We propose a fine-grained learning framework based a novel approach of joint step segmentation and key action scorer for hand hygiene assessment. 
    
    \item We design a multi-stage convolution-transformer network to establish the long-range dependence between frames without increasing too much computation for accurate step segmentation. It effectively alleviates the problem of over-segmentation in conventional multi-stage convolutional networks by embedding linear transformers.
    
    \item To accurately assess the quality of each step which involves several key actions, we design the key action scorer based on the multiple branch predictions, in which the Sigmoid functions are all learnable with independent parameters to model the characteristics of different key actions.
    
    \item We create the first unified and high-quality video dataset for hand hygiene assessment. It provides both video-level and frame-level annotations under the supervision of medical staff and thus would effectively promote the research and development of hand hygiene assessment. 
    
\end{itemize}


\section{Related Work}

 \subsection{Action Segmentation}
Early action segmentation methods use sliding windows combined with non-maximum suppression~\cite{Rohrbach} to focus on short-term dependence, but more windows lead to expensive calculation costs. Most recent methods use sequential convolution networks to pay attention to both short-term and long-term dependencies. MS-TCN~\cite{Yazan} uses the temporal convolutional networks to aggregate the temporal information and uses multi-step TCN to better adjust the classification results. 
Li et al.~\cite{Li} add the dilated convolution to solve the problem that some local information can not be extracted. BCN~\cite{Wang} uses the adaptive cascade network to distinguish difficult samples, which greatly improves the classification accuracy of difficult samples, and the temporal regularization method combined with action boundary information reduces the over-segmentation. The time receptive field of the model plays an important role in action segmentation. A large receptive field contributes to the long dependence between videos, while a small receptive field helps to capture local details. Gao et al.~\cite{Gao} propose to find a better combination of receptive fields through a global to local search scheme. Yi et al.~\cite{Yi} solve the problems encountered when introducing transformer into action segmentation task and design an efficient model based on transformer.

\subsection{Action Assessment}
As for hand hygiene assessment, Zhong et al.~\cite{zhong2021designing} apply an iterative engineering process to design a hand hygiene action detection system to improve food-handling safety, it uses the results of action classification to achieve hand hygiene assessment.
Research on hand hygiene is more about pose estimation and detection tasks, which are different from hand hygiene assessment.

There are many researches on general action quality assessment. Existing methods usually regard it as a regression task, and the ultimate goal is to narrow the gap between the regression score and the ground truth score given by experts. 
Pirsiavash et al.~\cite{Pirsiavash} use the discrete cosine transform to encode joint trajectory as input feature and the support vector regression to construct the mapping from feature to final score. 
Based on the attention mechanism used by human beings, when evaluating videos, Li et al.~\cite{Li1} propose a spatial attention model based on a recurrent neural network, which considers the accumulated attention state from previous frames and advanced knowledge about the progress of ongoing tasks. 
Parmar et al.~\cite{Parmar} prove that C3D~\cite{DuTran} can effectively preserve the spatio-temporal information in the video, and help to improve the performance of evaluation tasks. Pan et al.~\cite{Pan} use I3D~\cite{Carreira} as the backbone network to extract the spatio-temporal features, and establish a trainable joint diagram and analyze their joint movements. 
Xu et al.~\cite{xu2019learning} proposed a deep architecture that includes two LSTM to learn the different scale features of videos, and presented a figure skating sports video dataset.
Recently, Zeng et al.~\cite{Zeng} not only use the dynamic information in the video but also pay attention to the static pose information. By combining static and dynamic information in a graph-based context-aware attention module, the good video representation can be achieved. 
Using the idea of comparative learning, Yu et al.~\cite{Yu} reformulate the problem of action assessment as the regression relative score concerning another video.
Xu et al.~\cite{xu2022finediving} construct a new fine-grained dataset and propose a procedure-aware approach to quantify quality differences between query and exemplar in a fine-grained way for action assessment.
Jain et al.~\cite{jain2020action} proposed a action scoring system based deep metric learning that learns similarity between two videos.
To solve the difficulty of manually label data, Zhang et al.~\cite{zhang2022semi} explored semi-supervised action assessment based adversarial learning and recover masked segment feature of an unlabeled video.

\begin{figure*}[ht]
    \centering
    \includegraphics[width=1\linewidth,height=1.1\columnwidth]{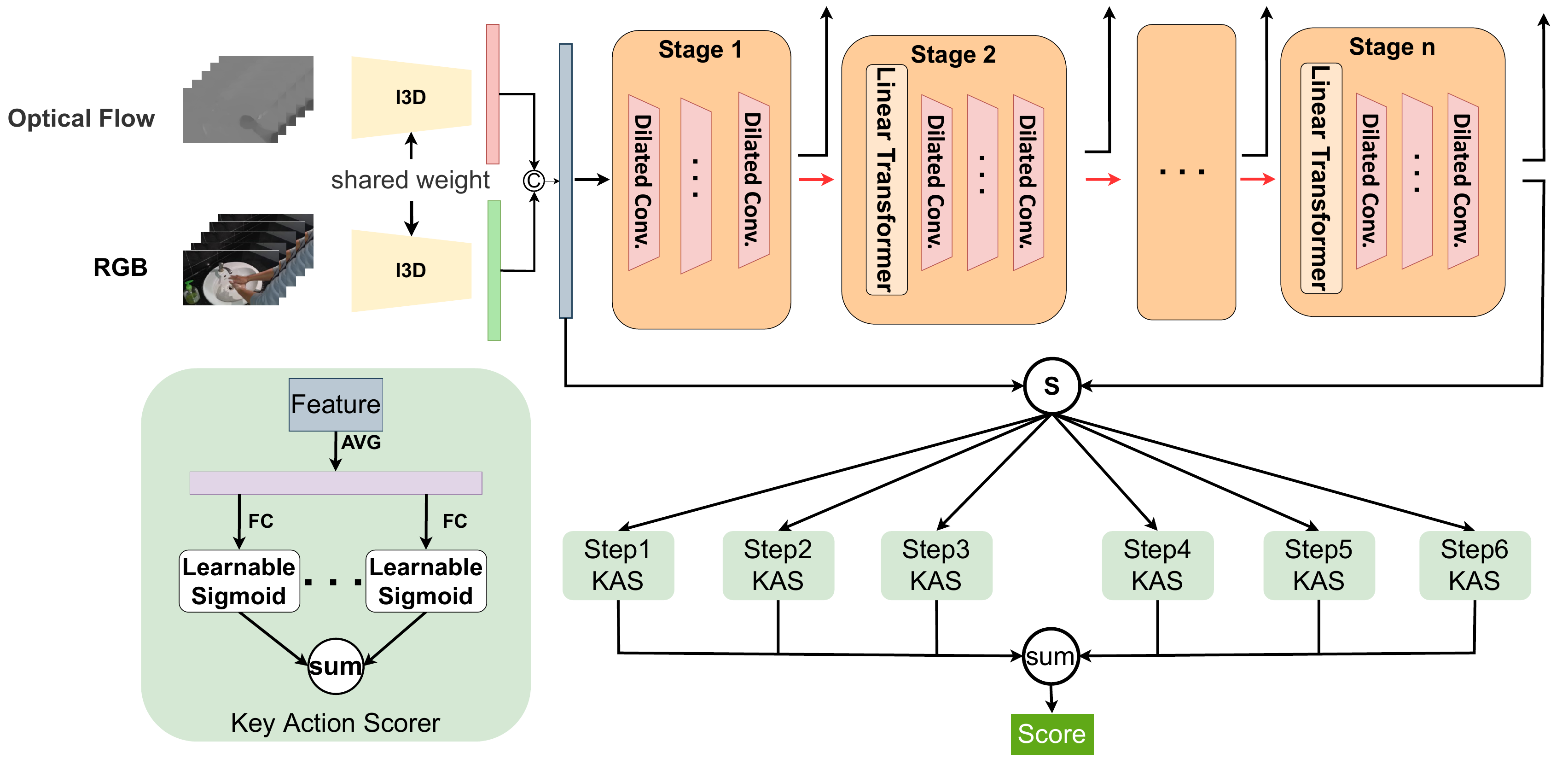}
    \caption{Pipeline of our framework. Herein, S indicates the selection of step features, sum stands for the summation, AVG and FC are the global average pooling and the full connection layer respectively.}
    \begin{picture}(0,0)
        \put(6,308){{$L_{seg}$}}
        \put(97,308){{$L_{seg}$}}
        \put(148,308){{$L_{seg}$}}
        \put(245,308){{$L_{seg}$}}

    \end{picture}
    \label{fig:framework}
\end{figure*}
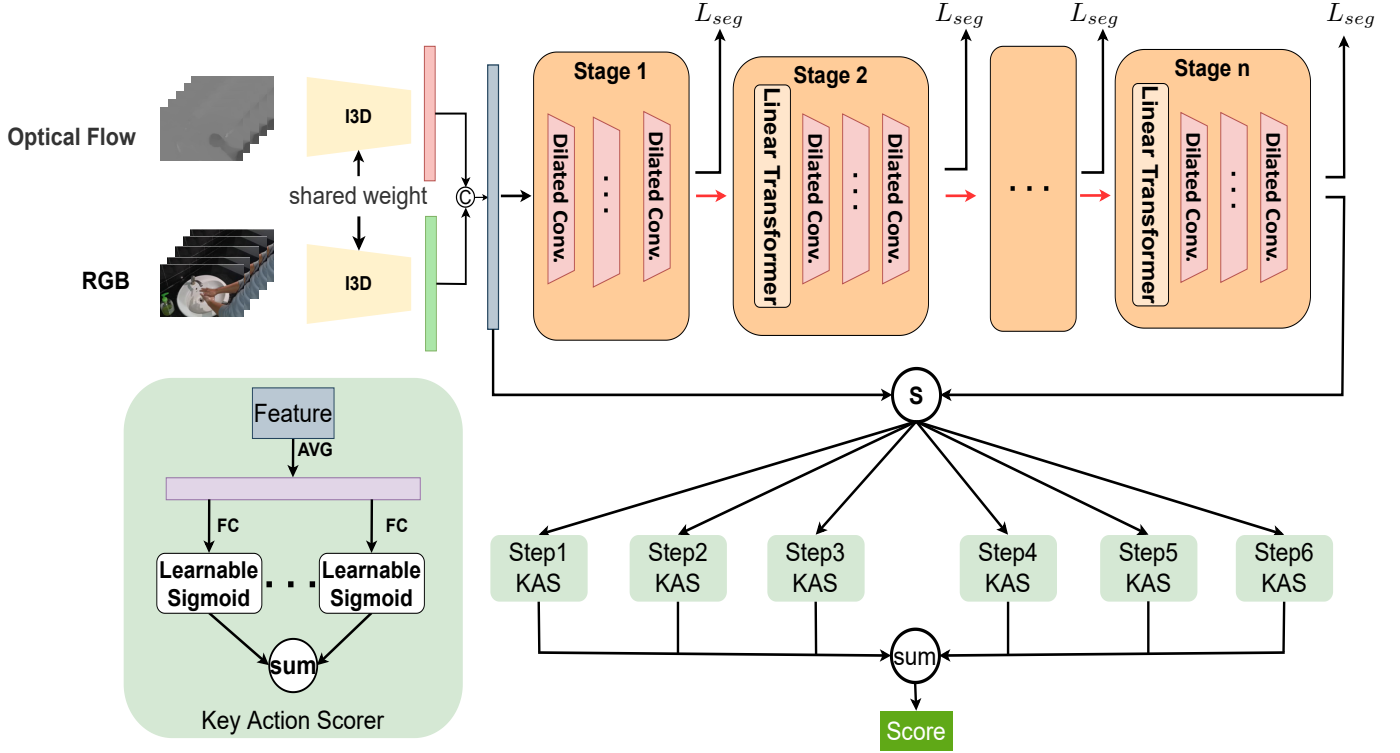

\section{Methodology}

This section will introduce the details of the proposed hand hygiene assessment approach via joint step segmentation and key action scorer, including video feature encoding, multi-stage convolution-transformer, key action scorer, and loss function. We first overview our framework for better readability.

\subsection{Overview}
The overall framework is shown in Fig.~\ref{fig:framework}.
Our framework for hand hygiene assessment mainly consists of two parts. The first part is the multi-stage convolution-transformer network for step segmentation , and the second part is the action assessment model based on key action scorer. 
First, the optical flow in hand hygiene video is extracted. The I3D feature extractor~\cite{Carreira} takes RGB and optical flow information as inputs to compute the corresponding appearance and motion features. Then, these two features are concatenated and inputted into the multi-stage convolution-transformer network. In the network, each stage produces corresponding features, and we use the linear transformer to build the long-range dependence and thus enhance these features. After that, we obtain the temporal segments corresponding to each step and evaluate these temporal segments by predicting the quality scores respectively by the key action scorers. Finally, we sum up the score of each step to obtain the final assessment score.

\subsection{Video Feature Encoding}
Existing action segmentation networks~\cite{Li,Wang} usually only use RGB features for subsequent prediction. While the optical flow features~\cite{zhao2017pooling, alwando2018cnn} are important for motion analysis including segmentation and assessment. On this basis, we further extract the optical flow features of video. 
Specifically, given an unprocessed hand hygiene video $X_{1:T} = \{x_t\}_{t=1}^T$ with $T$ frames, we first extract the optical flow to obtain the optical flow $F_{1:T} = \{f_t\}_{t=1}^T$, and then use the pre-trained feature extraction network I3D~\cite{Carreira} to encode RGB information and optical flow information respectively. Finally, we obtain two kinds of features 
$\Phi_{rgb} = \{\phi(x_1),\phi(x_1),...,\phi(x_T)\} \in \mathbb{R}^{T\times D}$
and       
$\Phi_{f} = \{\phi(f_1),\phi(f_1),...,\phi(f_T\}\in \mathbb{R}^{T\times D}$,
which represent RGB features and optical flow features respectively.


\subsection{Step Segmentation}

Most existing works~\cite{Yu,xu2022finediving} evaluate short action such as diving, where the input is a whole video containing the short action and the output is the score of the action. Specific, given the input video feature $\Phi = \{\phi(x_1),\phi(x_1),...,\phi(x_T)\} \in \mathbb{R}^{T\times D}$. These works formulate action assessment task as a regression problem is to predict the action score $S$:
\begin{equation}
S=R(\Phi),
\end{equation}
where $R$ is the regression model. 
However, these methods can not perform an accurate evaluation of long action because their quality depends on the completion quality of each step. Therefore, we propose to segment the steps before evaluating the action quality. 
Most of existing action segmentation methods~\cite{Yazan, Li} adopt multi-stage convolutional models. They predict a rough segmentation result in the first stage, and the result of the previous stage is gradually refined in each subsequent stage. However, the actions in the hand hygiene video are continuous and similar, which leads to segmentation errors (i.e., over-segmentation) only depending on short-range dependence in multi-stage convolutional models. 

To obtain the accurate results of step segmentation, we propose to embed the linear transformer~\cite{katharopoulos2020transformers} in the multi-stage convolution model~\cite{Wang} to form the multi-stage convolution-transformer network, which can model the long-range dependence between frames without increasing too much computation compared with the multi-stage convolutional model.

The transformer is first applied to the research of natural language processing~\cite{vaswani2017attention}. It uses a self-attention mechanism instead of the sequence structure of RNN so that the model can be trained in parallel and get global information. The input vectors of the self-attention mechanism are usually named query, key, and value. The weight distribution of the value vector is determined by the similarity between the query vector and the key vector. Formally, the attention layer is denoted as: 
\begin{equation}
V'=softmax(\frac {QK^T}{\sqrt{d_k}})V.
\end{equation}
where $Q$, $K$, and $V$ indicate the query, key, and value respectively and $d_k$ is the vector dimension.
To reduce the risk of over-segmentation caused by the lack of long-range dependence in multi-stage convolutional networks, we introduce the transformer to re-model the long-range dependence. Specifically, we design a multi-stage convolution-transformer network based step segmentation, and each stage includes several dilated convolutions with different dilated rates to establish the dependence of short-range frames. We first generate a rough result and a stage feature in the first stage using the feature extracted by I3D and then make further stage predictions. In each stage, we use the prediction result and stage feature of the previous stage to compute the refined prediction result and stage feature $\Phi_n\in\mathbb{R}^{T\times D_n}$, where $n$ indicates the number of stages. In addition, we use the stage feature as key, value, and query vectors at the same time, and enhance the feature of previous stage through self-attention, to establish the global dependency between all frames.

Note that the original self-attention mechanism is calculated by matrix multiplication. With the increase of sequence length $T$, the calculation cost is extremely expensive. Therefore, we introduce a variant of transformer, linear transformer~\cite{Katharopoulos}, which uses another kernel function to measure correlation instead of matrix multiplication to reduce the calculation amount of transformer and has similar performance. Specifically, we use 
\begin{equation}
sim (Q,K) = \theta(Q)\cdot \theta(K)^T.
\end{equation}
where $\theta(x) = elu(x) +1 $. $elu()$~\cite{Clevert} is an improved version of ReLU function~\cite{Glorot}. After this transformation, we only compute $\sum_{j=1}^{T}\theta(K_j)V^T_j$ and $\sum_{j=1}^{T}\theta(K_j)$ once and reuse them for every query, and the computational complexity is reduced from $O(T^2)$ to $O(T)$ as follows:
\begin{equation}
\begin{aligned}
V'_i &=\frac {\sum_{j=1}^{T}sim(Q_i,K_j)V_j}{\sum_{j=1}^{T}sim(Q_i,K_j)} \\
&=\frac {\theta(Q_i)^T \sum_{j=1}^{T}\theta(K_j)V^T_j}{\theta(Q_i)^T \sum_{j=1}^{T}\theta(K_j)}).
\end{aligned}
\end{equation}
where $i$, $j$ are the row indexes of $Q$ and $K$, $V$ respectively and $Q = W_{q}\Phi_n,K=W_{k}\Phi_n,V=W_{v}\Phi_n$. $W_q,W_k,W_v$  represent the learnable linear transformation matrices.

In addition, the attention mechanism of the transformer model does not contain the positional information. To solve this problem, the transformer adds an extra vector of positional encoding to the input of the encoder layer and decoder layer. But we think that because the video itself has temporal information, the positional encoding in the transformer will disturb the temporal information, and thus we do not use positional encoding in our method. 
We add the linear transformer to all the later stages, and use its powerful global modeling ability to enhance the long-range dependence of the feature of each stage, so as to solve the over-segmentation problem.
Finally, the step segmentation result is obtained by the last stage. 
        


\subsection{Hand Hygiene Assessment}

After step segmentation, for each step, we choose the continuous and longest part as the representative segment. On one hand, a step might appear many times in video sequence due to possible wrong segmentation and we choose the continuous frames to filter out some wrong segmentation frames. On the other hand, after training the step segmentation model, even if there is wrong segmentation, the longest part is still correctly segmented for most frames. Therefore, we use these continuous and longest parts as the input of the module of hand hygiene assessment. After obtaining each step segment, accurately evaluating each step is the most critical issue. We observe that each step involves two key actions, which determine the quality of this step. To accurately evaluate each step in hand hygiene, we design a Key Action Scorer (KAS) to assess the key action in each step. Each step segment will input to KAS to compute the corresponding assessment score.

\begin{figure*}[ht]
    \centering
    \includegraphics[width=1\linewidth,height=1\columnwidth]{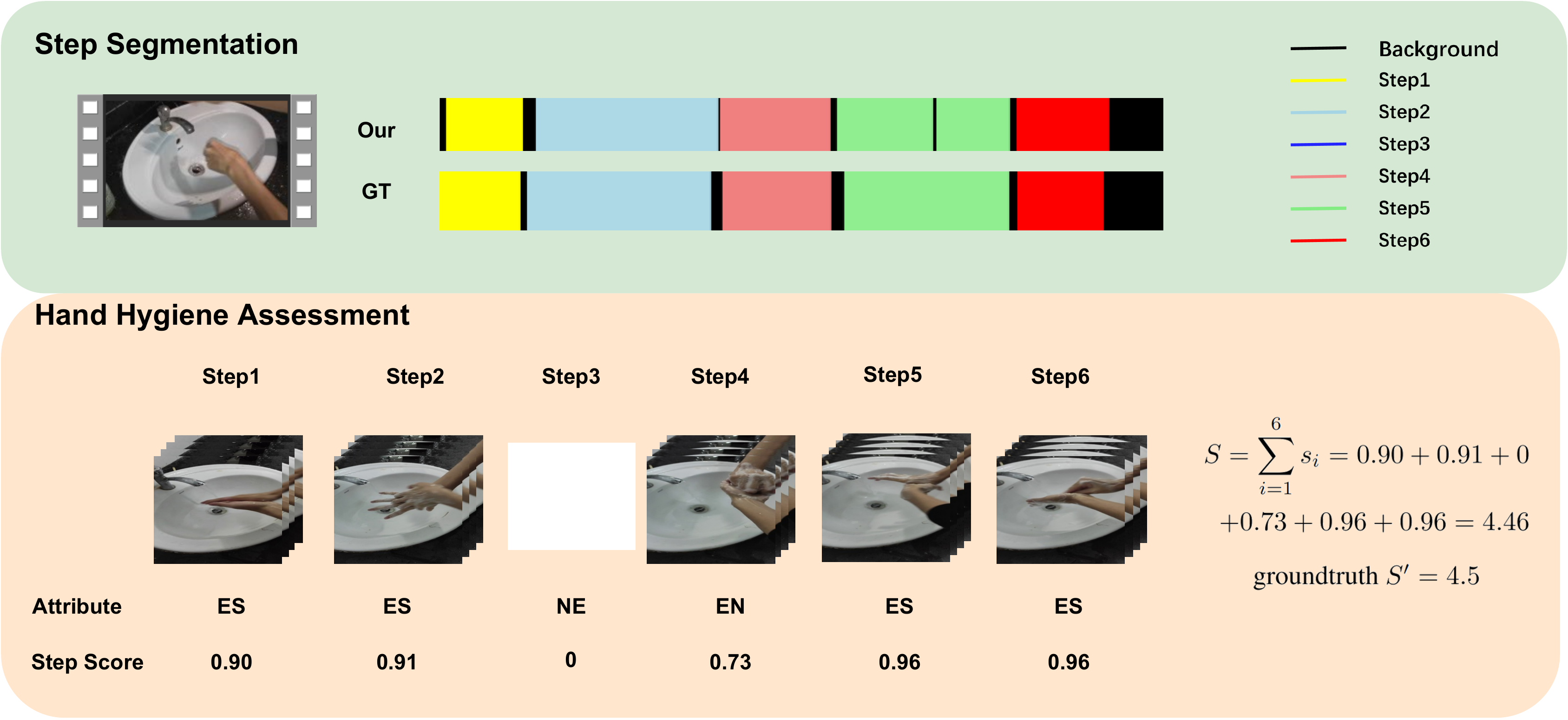}
    \caption{An example of step segmentation and assessment by our method. The step segmentation model divides a hand hygiene video into step-based video segments. The hand hygiene assessment model assesses the key actions of each step and finally computes the final assessment score of the whole video. Herein, Attribute is the attribute annotation of each step, and Step Score indicates the output of the key action scorer.}
    \label{fig:result}
\end{figure*}

\textbf{\flushleft Key Action Scorer.}
For each step, there are some key actions, which play a significant role in the standardization of this step. Therefore, it is necessary to accurately assess the quality of these key actions. 

Sigmoid, a common activation function, can map a variable between 0 and 1, which is widely used in different tasks. However, there are some differences among different key actions. Therefore, for each key action, a specific structure should be used to assess it. The standard Sigmoid is too steep and not suitable for the evaluation of different key actions.  While we introduce a learnable parameter into Sigmoid function to control the steepness, and use learnable Sigmoid for more accurate assessment of different key actions. The Sigmoid with learnable parameter is formulated as follows:
\begin{equation}
LS = \frac{1}{1+e^{-\lambda x}}
\end{equation}
where $x$ is the output feature of fully connected layers (FC), $\lambda$ is the learnable parameter that controls the steepness of Sigmoid function. 

In this work, based on the characteristics of the learnable Sigmoid function, we design a new hand hygiene assessment module including six Key Action Scorers (KAS). Each KAS corresponds to a hand hygiene step and consists of two branches with the same structure.  In particular, each branch includes a FC and a learnable Sigmoid, and different branches have independent parameters to model the characteristics of different key actions.
As shown in Fig.~\ref{fig:framework}, we first use global average pooling to pool the segment features and then input them into different key action assessment branches which are composed of FC and learnable Sigmoid. The number of key action assessment branches is determined by the number of key actions in each step. In our work, we set the number of branches is two, as shown in Table~\ref{tab:keyactionofstep}. We hope that each evaluation branch can learn a set of specific parameters to accurately evaluate specific key actions as follows:
\begin{equation}
s_i=\frac{1}{k}\sum_{j=1}^{k}LS_j(FC_j(AVG(\bm{\phi_i})))
\end{equation}
where $s_i$, $\bm{\phi_i}$ are the $i$-th step score and the segment feature respectively. $k$ is the number of the key action. $LS$, $FC$, and $AVG$ indicate the operations of learnable Sigmoid, fully connected layer, and global average pooling respectively. The final total score is the summation of all step scores as follows:
\begin{equation}
S = \sum_{i=1}^{6}s_i.
\end{equation}
where $S$ represents the total score of the hand hygiene video.

\subsection{Loss Function}
Our loss function is composed of two parts. The first part is the loss in the step segmentation model. It includes the cross entropy loss for frame-wise classification and logarithmic probability smoothing of censored mean square error, as follows:
\begin{equation}
{L}_{CLS}=\frac{1}{T} \sum_{t}-\log \left(y_{t, c}\right),
\end{equation}
\begin{equation}
\mathcal{L}_{T-MSE}=\frac{1}{T C} \sum_{t, c} \tilde{\Delta}_{t, c}^{2},
\end{equation}
\begin{equation}
\tilde{\Delta}_{t, c}= \begin{cases}\Delta_{t, c} & : \Delta_{t, c} \leq \tau \\ \tau & : \text { otherwise }\end{cases},
\end{equation}
\begin{equation}
\Delta_{t, c}=\left|\log y_{t, c}-\log y_{t-1, c}\right|,
\end{equation}
where $T$ is the video length, C is the number of calsser,and $y_{t, c}$ is the probability of class $c$ at time $t$. The the loss in the step segmentation model as follow:
\begin{equation}
\mathcal{L}_{SEG}=\mathcal{L}_{CLS}+\lambda \mathcal{L}_{T-M S E},
\end{equation}
please refer to~\cite{Yazan} for details.

The second part is the loss of the hand hygiene assessment, we use the mean square error to measure the difference between the predicted score and the ground truth score as follows:
\begin{equation}
\mathcal{L}_{MSE}=\frac{1}{n} \sum_{i=1}^{n}\left(\hat{y}_i-{y_i}\right)^{2}
\end{equation}
where $\hat{y}$ and $y$ represent the predicted score and the ground truth score, respectively.
The final loss function is a combination of the above mentioned losses:
\begin{equation}
\mathcal{L} = \mathcal{L}_{SEG} + \mathcal{L}_{MSE}
\end{equation}

The training of the step segmentation network and the action assessment network is formulated as a multi-task learning problem and performed in an end-to-end manner. In this way, the performance of both step segmentation and action assessment is boosted. We show a typical process of our method in Fig. \ref{fig:result}.

\section{HHA300: Hand Hygiene Assessment Dataset}
\begin{figure}[h]
    \centering
    \includegraphics[width=0.8\linewidth]{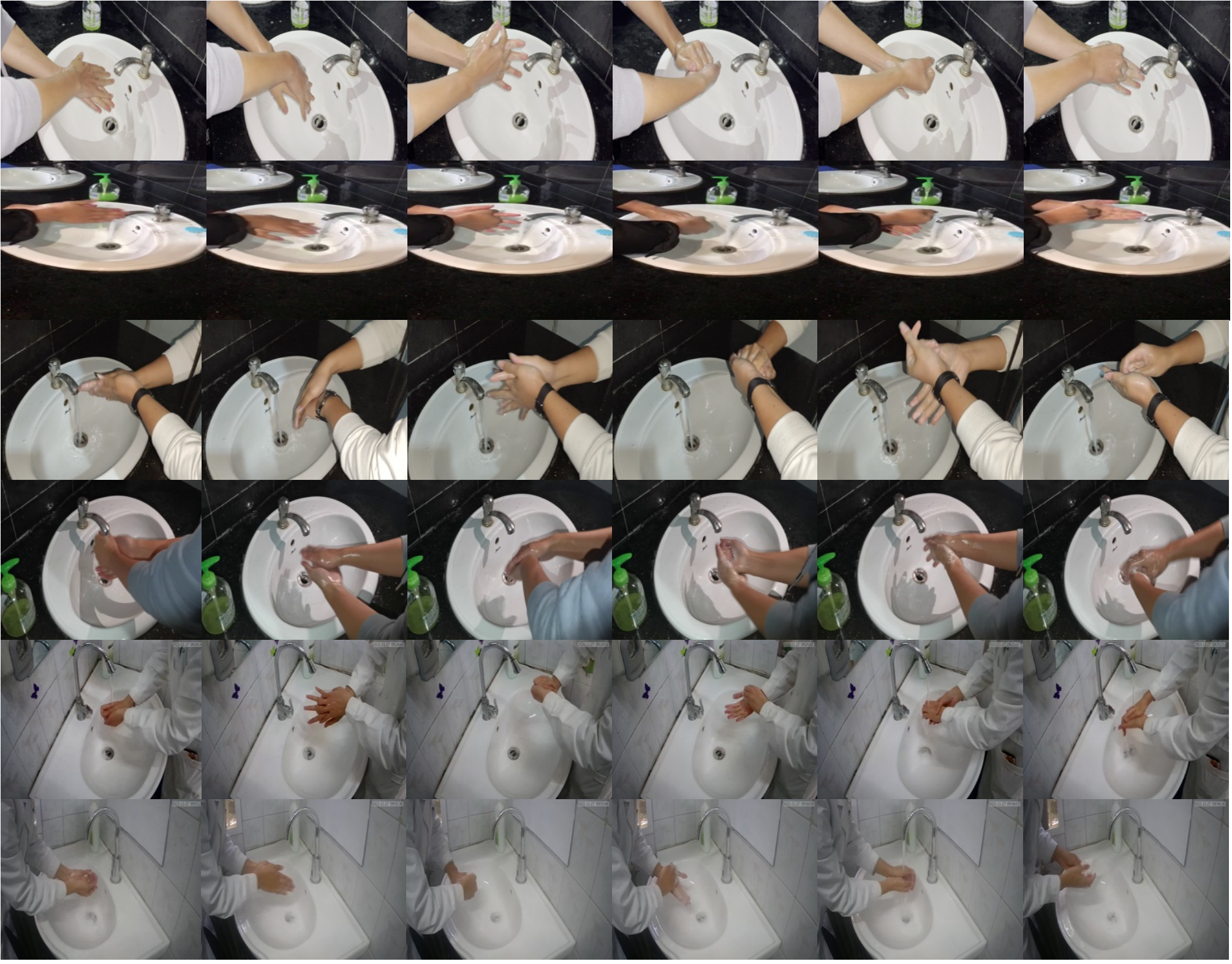}
    \caption{Data samples from different viewpoints and persons in HHA300 dataset. }
    \label{fig:dataset}
\end{figure}

To promote the research and development of hand hygiene assessment, we create a unified dataset for hand hygiene assessment, namely HHA300. 
we divide HHA300 into a training set and a testing set for facilitating the evaluation of step segmentation and action assessment models, as shown in Table \ref{tab:details of hand hygiene}. We also present some samples of HHA300 are showed in Fig. \ref{fig:dataset}. In this section, we introduce the details of HHA300.

\begin{table}[h]
  \centering
  \caption{The details of our HHA300 Dataset.}\footnotesize
  \begin{tabular}{@{}lccccc@{}}
    \toprule
     & Video & \begin{tabular}[c]{@{}c@{}}Min \\ frames\end{tabular} & \begin{tabular}[c]{@{}c@{}}Mean\\ frames\end{tabular} & \begin{tabular}[c]{@{}c@{}}Max \\ frames\end{tabular} & \begin{tabular}[c]{@{}c@{}}Total \\ frames\end{tabular}   \\
    \midrule
    {Train set}   &225        & 373          & 1048           & 1579       &   236k  \\
    {Test set}    & 75        & 406           &1026            &1436        &  77k   \\
    \bottomrule
  \end{tabular}
  \label{tab:details of hand hygiene}
\end{table}

\subsection{Data Collection}
Existing hand hygiene datasets~\cite{vo2021fine,ivanovs2020automated} are all used for hand hygiene image classification. There lacks a dataset for hand hygiene assessment. To handle this problem, we create a unified hand hygiene assessment dataset HHA300, under the supervision of medical staff. In specific, we use CCD cameras to capture video data in various scenes and invite 60 persons (including ordinary people and professional medical staff) to divide into several groups to conduct hand hygiene behaviors with different standards from different scenes. Our dataset contains all most possible situations in the real world. In addition, we also invite several medical staff for professional guidance in data creation. In this way, we collect a total of 300 hand hygiene video sequences with an average video length of over 1000 frames. 

\subsection{Fine-grained Annotation}
To provide high-quality fine-grained annotation, in addition to a total assessment score for each video, we also annotate the frame-level labels under the supervision of medical staff.
An action segmentation and action assessment dataset needs to have high-quality frame class annotation and score annotation, which are essential for training a robust model and ensuring the fairness of performance evaluation. Therefore, we make professional annotations with the help of medical staffs, and every frame is checked to ensure the accuracy of frame-level annotation.
In specific, according to the hand hygiene standard of WHO, hand hygiene is divided into six steps. We divide all frames into 7 categories, and they are palm to palm (step 1), palm over dorsum with fingers interlaced (step 2), palm to palm with fingers interlaced (step 3), back of fingers to opposing palm (step 4), rotational rubbing of the thumb (step 5), fingertips to palm (step 6) and background action. We also annotate a total assessment score for each video. To ensure high-quality annotations, we establish a series of evaluation criteria under the help of medical staffs to assess each video sequence uniformly and fairly.

\subsection{Challenge}
\label{attributes}
The complexity and type of a scene are key factors in enhancing the diversity of the dataset. To this end, We collect videos in HHA300 from a wide range of people, camera viewpoints, scene complexity, and other environmental factors. To clarify the advantages of HHA300, we analyze its diversity from the following aspects.
First of all, unlike other existing datasets~\cite{vo2021fine,ivanovs2020automated}, the shooting scenes are very rich, including the infirmary, public toilet, and dormitory. Secondly, we have a large number of photographers, including dozens of medical staff and students. Besides, the shooting angle is not fixed, and our dataset has six different camera viewpoints, such as top, left, and right. These factors are with different complexities and thus bring some difficulties for step segmentation and hand hygiene assessment. 

\begin{figure}[h]
    \centering
    \includegraphics[width=1\linewidth]{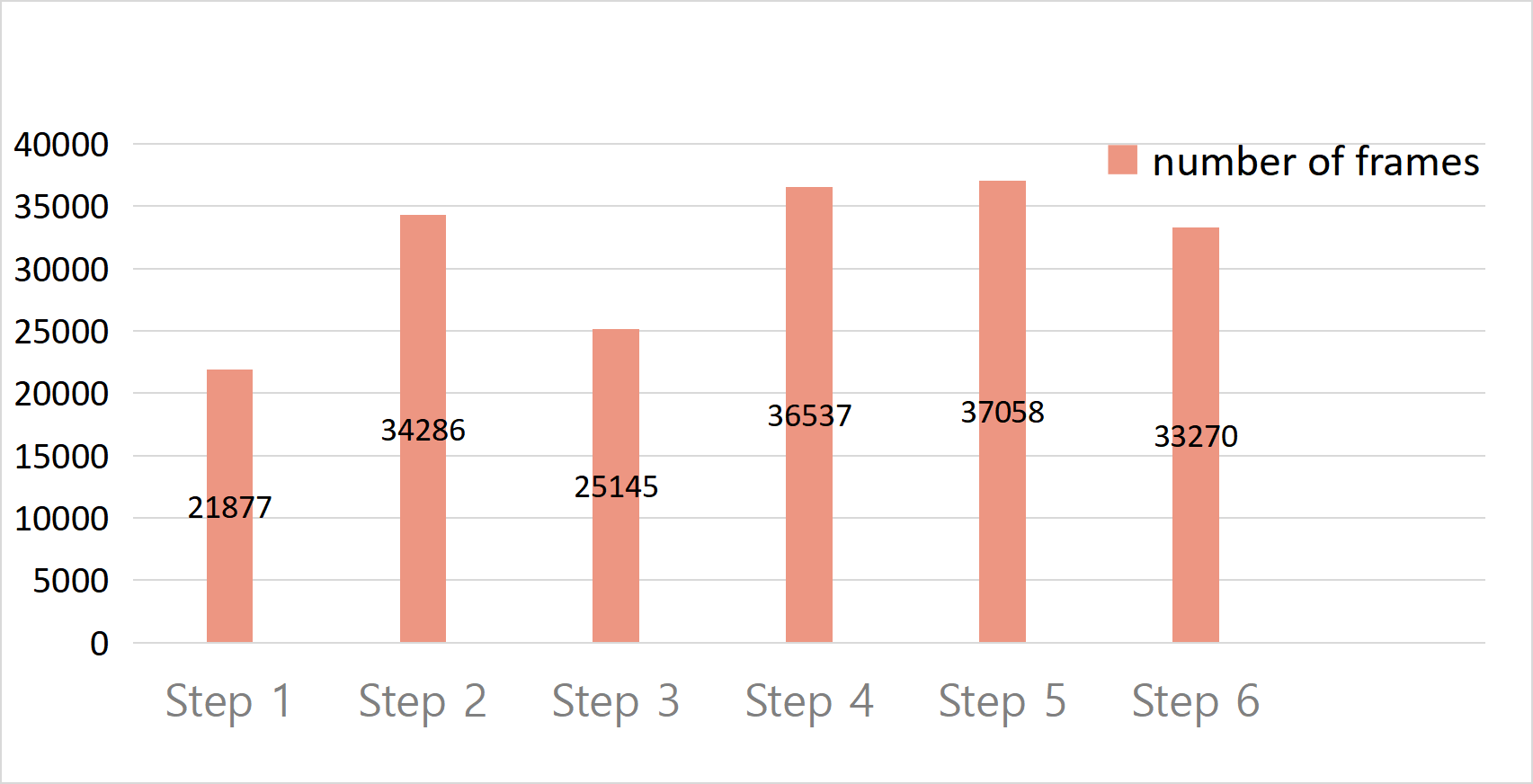}
    \caption{The number of frames per step in HHA300 dataset. }
    \label{fig:frame number}
\end{figure}

In addition to the above challenges caused by external factors, we are also considering the challenges caused by internal factors. In the task of hand hygiene assessment, the action quality of each step has a certain influence on the result of the assessment model. Therefore, we simulate more real-world situations in data creation. 
In specific, for each step, we define three degrees of standardization as follows. 1) The step does not exist (NE); 2) The step exists but is nonstandard action (EN); 3) The step exists and is standard (ES). Table \ref{tab:dis of hand hygiene} shows the video distribution of attributes on our dataset. We ask people to do hand hygiene to consider different degrees of standardization in each step. In this way, our HHA300 contains almost all possible real-world challenging situations. 

\begin{table}[h]
  \centering
   \caption{Video distribution on different attributes in HHA300 dataset.}\footnotesize
  \begin{tabular}{lcccccc}
    \toprule
    \diagbox{Attributes}{Number}{Step} & 1   & 2 & 3  &4 &5 &6 \\
    \midrule
    {NE} &37     & 46    &48   &49   &42    & 53                      \\
    \midrule
    {EN} & 27    & 25    &32   &34   &21    & 38                      \\
    \midrule
    {ES} & 163   & 156    &147   &144   &164    & 136                      \\
    \bottomrule
  \end{tabular}
  \label{tab:dis of hand hygiene}
\end{table}

\section{Experiments}

\subsection{Evaluation Setting}
In this section, we will describe implementation details, evaluation metrics, datasets and evaluation methods in our experiments.
\textbf{\flushleft Implementation.} 
For our hand hygiene dataset HHA300, we first extract the optical flow from hand hygiene videos. Then, all frames are resized to 224×224. Finally, we extract 1024-dimensional RGB and optical flow features from the I3D model pretrained on Kinetics~\cite{Carreira}. In all experiments, we use Adam optimizer with a learning rate of 0.0005, and the weight decay is set to 0.

\textbf{\flushleft Evaluation Metrics.}
For the evaluation of step segmentation, we employ the frame-wise accuracy (Acc), segmental edit distance, and the segmental F1 scores ~\cite{lea2017temporal}at overlapping thresholds 10\%, 25\%and 50\%, denoted by F1@\{10,25,50\}. 
As we all know, Acc is one of the most common evaluation metrics of step segmentation. since Acc is related to the number of frames, the influence of the long action class on this metric is greater than that of the short action class, which makes this metric unable to reflect the error of over-segmentation. 
To this end, two metrics are introduced, including the segmental edit distance and the segmental F1 score, which can penalize over-segmentation errors. The edit distance penalizes over-segmentation errors by predicting the order of actions, while the F1 score is similar to mean average precision (mAP) in detection task.

For the evaluation of hand hygiene assessment, similar to existing works~\cite{Yu,Zeng}, we use the standard evaluation metric known as Spearman’s rank correlation coefficient $\rho$.
which is defined as:
\begin{equation}
\rho = \frac {\sum_{i}(p_i-p)(q_i-q)}{\sqrt{\sum_{i}(p_i-\bar{p})\sum_{i}(q_i-\bar{q})}}
\end{equation}
where $p$ and $q$ represent the ranking for each sample of two series respectively. We also use the relative L2-distance(R-$\ell_2$)~\cite{Yu} which is defined as:
\begin{equation}
\mathrm{R}-\ell_{2}(\theta)=\frac{1}{K} \sum_{k=1}^{K}\left(\frac{\left|s_{k}-\hat{s}_{k}\right|}{s_{\max }-s_{\min }}\right)^{2}
\end{equation}
where $s_{min}$ and $s_{max}$ are the highest and lowest scores for an action, $s_k$ and $\hat{s}$ represent the ground truth score and the prediction score for the $k$-th sample, respectively. Spearman’s correlation focuses more on the ranks of the predicted scores while the R-$\ell_2$ focuses on the numerical values of the predicted scores.

\textbf{\flushleft Datasets.}
We evaluate the proposed method on our HHA300, and further evaluate action segmentation methods on three public datasets, including \textbf{50Salads}~\cite{Stein}, \textbf{GTEA} ~\cite{Fathi} and \textbf{Breakfast} ~\cite{Kuehne} datasets. Tab \ref{tab:compare with action seg} shows the comparison of these datasets.

\begin{table*}[h]
  \centering
  \caption{Comparing with public action segmentation datasets.}\footnotesize
  \begin{tabular}{@{}lccccccc@{}}
    \hline
    Datasets              & Videos  & Action classes &View   &Description\\
    \hline
    {50Salads}~\cite{Stein}   &50 &17 &Top-view &Salad preparation activities \\
    
    {GTEA}~\cite{Fathi}   &28 &11 &Egocentric &7 Different activities, like preparing coffee or cheese sandwich \\
    {Breakfast} ~\cite{Kuehne}   &1712 &48 & Third person view & Breakfast preparation related activities in 18 different kitchens \\
    \hline
    {HHA300(our)} &300 & 7 & Third person view &Hand hygiene behavior of different people in different scenes     \\
    \hline
  \end{tabular}
  \label{tab:compare with action seg}
\end{table*}

The \textbf{50Salads}~\cite{Stein} dataset contains 50 videos with 17 action classes. As the name of the dataset indicates, the videos depict salad preparation activities. For evaluation, we use five-fold cross validation and report the average result. 

The \textbf{Georgia Tech Egocentric Activities (GTEA)} ~\cite{Fathi} dataset contains 28 videos corresponding to 7 different activities, like preparing coffee or cheese sandwich, and is performed by 4 subjects. The frames of the videos are annotated with 11 action classes including background. On average, each video has 20 action instances. We use cross-validation for evaluation and report the average. 

The \textbf{Breakfast} ~\cite{Kuehne} dataset is the largest among the three datasets with 1,712 videos. The videos are recorded in 18 different kitchens showing breakfast preparation related activities. For evaluation, we use the standard 4 splits as proposed in~\cite{Kuehne} and report the average result.


\textbf{\flushleft Evaluation Methods.}
To facilitate comparison, we evaluate some step segmentation models on HHA300 dataset. In addition, following the settings for competitors given in ~\cite{xu2019learning}, we also evaluated some action assessment methods. we consider different combinations of the following model components.
\begin{itemize}
    \item \textbf{Feature extractor} We use three kinds of pretrained feature extractors, namely ResNet50~\cite{He}, I3D~\cite{Carreira}, C3D~\cite{DuTran}. For I3D and C3D, we extract RGB features and optical flow features separately. For ResNet50, we use the outputs of the fifth layer for RGB images and optical flow, and then the average pooling to obtain the features with same dimension.
    
    \item \textbf{MLP} We first use either average or maximum pooling for video-level description and then predict the score through a two-layer MLP, which is optimized by the MSE loss between the prediction and the ground truth.
    
    \item \textbf{LSTM~\cite{hochreiter1997long}} Similar to Parmar et al.~\cite{Parmar}, we generate video-level descriptions by a LSTM architecture. In our setting, the hidden dimensions of the LSTM layers is 256, and one fully connected layer for regression.
\end{itemize}
In addition to the combination of feature extractor, MLP and LSTM, we also combine the advanced step segmentation model with MLP such as MS-TCN~\cite{Yazan}-MLP.

\subsection{Analysis}
\textbf{\flushleft Hand Hygiene Assessment}.
We evaluate our framework against some methods on HHA300 dataset, as shown in Table ~\ref{tab:ass on hygiene}. From the results, we can see that for the combinations of different feature extractors, MLP and LSTM, two methods based on LSTM is better than other methods based on MLP. LSTM can perform better feature descriptions by enhancing long-range dependency. But the evaluation results on all these methods are not good. The main reason is that these methods take all features of a video as the inputs of evaluation models to regress an assessment score, which does not consider fine-grained hand hygiene actions and irrelevant background actions.

For the methods based on step segmentation models and MLP, the evaluation results are better than the above methods. It is mainly because action segmentation models can divide hand hygiene into different steps for fine-grained assessment. Among these methods, BCN~\cite{Wang}-MLP has the best performance, mainly because BCN with boundary-aware cascade can perform more accurate step segmentation. Therefore, the performance of step segmentation is critical to hand hygiene assessment.

From the results, we can see that our framework significantly outperforms other methods in both two metrics. In particular, our method achieves 0.842 in Spearman’s rank correlation coefficient and 0.85 in relative L2-distance. Compared with other methods, our framework includes both the multi-stage convolution-transformer network for step segmentation and the key action scorer for each step of hand hygiene assessment. It not only segments long actions in hand hygiene videos into step-based segments but also makes an accurate assessment based on key actions for each step.   

\begin{table*}
  \centering
    \caption{Results of step segmentation and hand hygiene assessment on HHA300 dataset. Avg stands for the average pooling, Max stands for the maximum pooling, MLP refers to a multi-layer perceptron with two hidden layers, and LSTM is the long short-term memory.}
  \begin{tabular}{@{}lccccc@{}}
    \toprule
    Method               & F1@\{10,25,50\}$\uparrow$    & Edit$\uparrow$    & Acc$\uparrow$ & Spearman’s Correlation$\uparrow$         &$\mathrm{R}-\ell_{2}(*100)\downarrow$\\
    \midrule
    {ResNet50-Avg-MLP}  &- &- &- & 0.245                         & 39.92   \\
    {ResNet50-Max-MLP}  &- &- &- & 0.281                         & 37.22      \\
    {ResNet50-LSTM}     &- &- &- & 0.311                         & 36.97       \\
    \midrule
    {C3D-Avg-MLP}       &- &- &- & 0.274                        & 37.97     \\
    {C3D-Max-MLP}       &- &- &- & 0.286                         &  38.54         \\
    {C3D-LSTM}          &- &- &- & 0.350                         & 36.81        \\
    \midrule
    {I3D-Avg-MLP}       &- &- &- & 0.378                         & 36.50     \\
    {I3D-Max-MLP}       &- &- &- & 0.389                        & 36.13            \\
    {I3D-LSTM}          &- &- &- & 0.406                         & 34.60        \\
    \midrule
    {MS-TCN~\cite{Yazan}-MLP}         & 82.0 81.7 75.6     & 74.6    & 88.7 &0.704 &2.52\\
    {MS-TCN++~\cite{Li}-MLP}       & 83.3 83.3 75.9     &77.9 &89.0 &0.711 &2.01\\
    {ASFR~\cite{ishikawa2021alleviating}-MLP} &88.2 88.7 82.4 &82.0 &\textbf{89.9} &0.728 &1.80   \\
    {BCN~\cite{Wang}-MLP}  &87.4 87.1 81.1 &81.3 &89.1 &0.774 &1.58\\
    
    \midrule
    {Ours}          & \textbf{89.7 89.2 83.0} & \textbf{83.3}  & 89.1 & \textbf{0.852}                 & \textbf{1.07}     \\

    \bottomrule
  \end{tabular}

  \label{tab:ass on hygiene}
\end{table*}

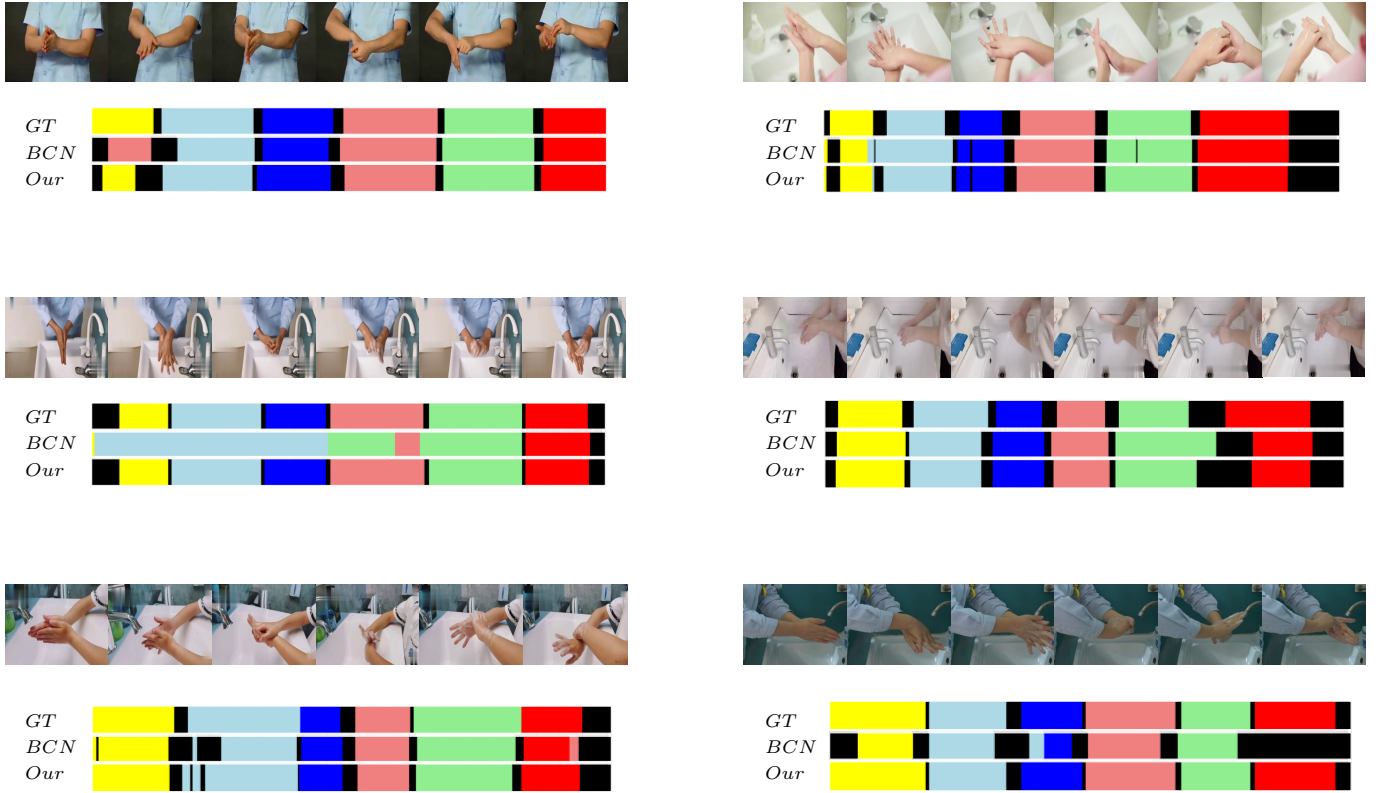
\begin{figure*}[ht]
    \centering
    \includegraphics[width=1\linewidth,height=1.2\columnwidth]{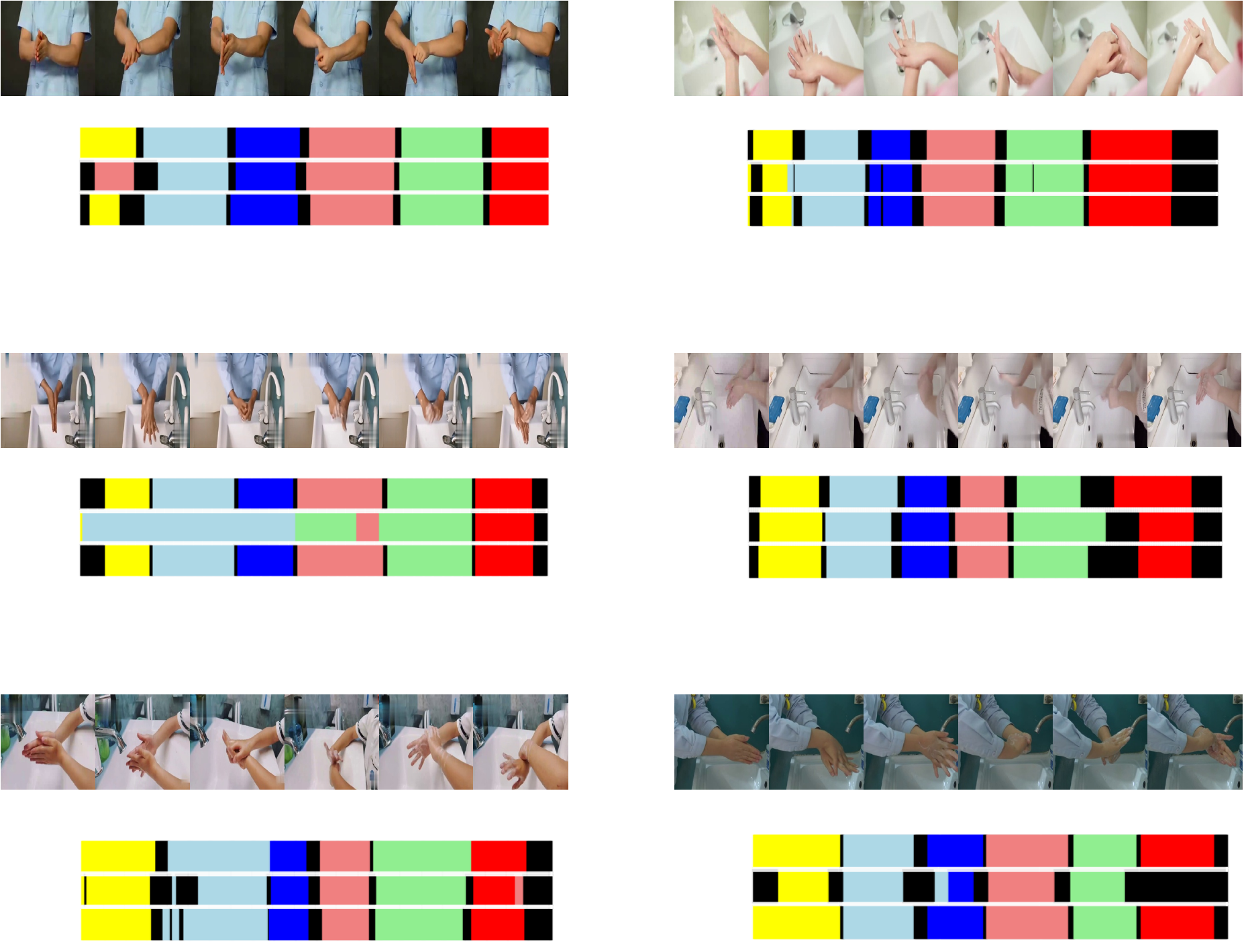}
    \begin{picture}(0,0)
        \put(-250,265){\scriptsize{$GT$}}
        \put(-250,255){\scriptsize{$BCN$}}
        \put(-250,245){\scriptsize{$Our$}}
        
        \put(30,265){\scriptsize{$GT$}}
        \put(30,255){\scriptsize{$BCN$}}
        \put(30,245){\scriptsize{$Our$}}
        
        \put(-250,155){\scriptsize{$GT$}}
        \put(-250,145){\scriptsize{$BCN$}}
        \put(-250,135){\scriptsize{$Our$}}
        
        \put(30,155){\scriptsize{$GT$}}
        \put(30,145){\scriptsize{$BCN$}}
        \put(30,135){\scriptsize{$Our$}}
        
        \put(30,40){\scriptsize{$GT$}}
        \put(30,30){\scriptsize{$BCN$}}
        \put(30,20){\scriptsize{$Our$}}
        
        \put(-250,40){\scriptsize{$GT$}}
        \put(-250,30){\scriptsize{$BCN$}}
        \put(-250,20){\scriptsize{$Our$}}
        
    \end{picture}
    \caption{Qualitative results of step segmentation on some samples from HHA300 dataset.}
    \label{fig:via}
\end{figure*}

\textbf{\flushleft Step Segmentation}.
For the multi-stage convolution-transformer based step segmentation, we evaluate it on HHA300 dataset shown in Table \ref{tab:ass on hygiene} and compare it with four multi-stage networks. As can be seen from the results, our method is superior to other methods in three metrics. Although we are only 0.2\% higher than BCN in accuracy, we are 3.7\% higher in segmental edit distance and \{2.3\%, 3.4\%, 2.7\% \} higher in the segmental F1 scores. The accuracy is related to the number of frames which makes this metric unable to reflect the error of over-segmentation, while the segmental edit distance and the segmental F1 scores can well reflect the ability of the model to reduce over-segmentation. Therefore, compared with other multi-stage convolution models, the experimental results show that our multi-stage convolution-transformer network can effectively alleviate over-segmentation while ensuring accuracy. Qualitative results on HHA300 dataset are shown in Fig. ~\ref{fig:via}. Predictions of baseline model BCN~\cite{Wang} have some over-segmentation errors, but our framework can reduce these errors by the multi-stage convolution-transformer network.

In addition, we evaluate our method on three challenging datasets, including 50Salads, Georgia Tech Egocentric Activities (GTEA), and Breakfast datasets. The results are shown in Table~\ref{tab:seg of 50Salads}. It can be seen that we achieve the state-of-the-art performance in terms of frame-wise accuracy on all datasets and competitive segmental edit distance and segmental F1 score. Compare with the baseline model BCN, we achieve 0.8\%, 1.5\%, and 1.1\% improvements in frame-wise accuracy on 50Salads, GTEA, and Breakfast datasets respectively. We also achieve improvements in segmental edit distance and the segmental F1 score. On Breakfast dataset, our framework and BCN have similar performance in terms of segmental edit distance, but our framework outperforms it in F1@{10, 25, 50} by \{1.2\%, 1.6\%, 2.0\% \}. These results show that our framework can identify the action segments that overlap with the real segments and the ground truth. However, compared with ASRF~\cite{ishikawa2021alleviating}, although our framework is better than it in terms of frame-wise accuracy, the segmental edit distance and segmental F1 score of our framework are inferior to that of ASRF. Ishikawa et al.~\cite{ishikawa2021alleviating} think that correcting the prediction result by detecting the action boundary can improve segmental edit distance and segmental F1 score, but ASRF predicts action boundaries off by some margin, therefore affecting the frame-wise accuracy. Our framework can achieve a good balance in three metrics and improve other metrics without losing frame-wise accuracy.

\begin{table*}[ht]
  \centering
  \caption{Step segmentation results of our method against other methods on public datasets including 50Salads, GTEA and Breakfast.}\footnotesize
  \begin{tabular}{@{}l|c|c|c|c|c|c|c|c|c@{}}
    \hline
                    & \multicolumn{3}{c|}{\textbf{50Salads}} 
                    & \multicolumn{3}{c|}{\textbf{GTEA}} 
                    & \multicolumn{3}{c}{\textbf{Breakfast}} \\
    \hline
                   & Acc$\uparrow$ & Edit$\uparrow$ & F1@\{10,25,50\}$\uparrow$
                   & Acc$\uparrow$ & Edit$\uparrow$ & F1@\{10,25,50\}$\uparrow$
                   & Acc$\uparrow$ & Edit$\uparrow$ & F1@\{10,25,50\}$\uparrow$         \\
    \hline
    {IDT+LM}~\cite{Richard} &48.7 &45.8 &44.4 38.9 27.8
                                        & -    & -     & -         
                                        & -    & -     & -         \\
                                        
    {ST-CNN}~\cite{lea2016segmental} &59.4 &45.9 &55.9 45.6 37.1
                                        & 60.6    & -     & 58.7 54.4 41.9         
                                        & -    & -     & -         \\
    
    {Bi-LSTM}~\cite{singh} & 55.7 & 55.6 & 62.6 58.3 47.0  
                        & 55.5 & - & 66.5  59.0 43.6 
                        & -    & -     & -         \\
    
    {ED-TCN}~\cite{lea2017temporal} & 64.7 & 59.8 & 68.0 63.9 52.6  
                        & 64.0 & - & 72.2 69.3 56.0  
                        & 43.3    & -     & -         \\
                        
    {TDRN~\cite{lei2018temporal}}              & 68.1 & 66.0 & 72.9  68.5 57.2  
                        & 70.1 & 74.1 & 79.2 74.4 62.7  
                        & -    & -     & -         \\
                        
    {SSA-GAN}~\cite{gammulle2020fine}           & 73.3 & 69.8 & 74.9 71.7 67.0  
                        & 74.4 & 76.0 & 80.6 79.1 74.2  
                        & -    & -     & -         \\    
                        
    {MS-TCN}~\cite{Yazan}            & 80.7 & 67.9 & 76.3 74.0 64.5
                        & 76.3 & 79.0 & 85.8 83.4 69.8
                        & 66.3 & 61.7 & 52.6 48.1 37.9\\
                        
    {MS-TCN++}~\cite{Li}          & 83.7 & 74.3 & 80.7 78.5 70.1
                        & 80.1 & 83.5 & 88.8 85.7 76.0
                        & 67.6 & 65.6 & 64.1 58.6 45.9\\
                        
    {ASRF}~\cite{ishikawa2021alleviating}          & 84.5 & \textbf{79.3} & \textbf{84.9 83.5 77.3}
                        & 77.3 & 83.7 & \textbf{89.4} 87.8 \textbf{79.8}
                        & 67.6 & \textbf{72.4} & \textbf{74.3 68.9} 56.1\\

    {BCN}~\cite{Wang}     & 84.4 & 74.3 & 82.3 81.3 74.0
                        & 79.8 & 84.4 & 88.5 87.1 77.3
                        & 70.4 & 66.2 & 68.7 65.5 55.0\\

    \hline                    
    {Our}      & \textbf{85.2} & 76.5 & 83.7  82.6  76.3
                    & \textbf{81.3} & \textbf{84.7} & 89.0  \textbf{88.2}  78.0
                    & \textbf{71.5} & 67.7 & 69.9 67.1 \textbf{57.0}\\
    \hline
  \end{tabular}
  \label{tab:seg of 50Salads}
\end{table*}

\begin{table}[h]
  \centering
  \caption{Results of our method with and without optical flow features in step segmentation on HHA300 dataset.}\footnotesize
  \begin{tabular}{@{}lccccc@{}}
    \toprule
    Method              & Acc$\uparrow$    & Edit$\uparrow$    & F1@\{10,25,50\} $\uparrow$                 \\
    \midrule
    {Our$_{rgb}$} &88.2 & 81.7    &86.3 85.4 79.1                       \\
    \midrule
    {Our}         & \textbf{89.1} & \textbf{83.3} & \textbf{89.7 89.2 83.0}       \\
    \bottomrule
  \end{tabular}
  \label{tab:optical flow of hand hygiene}
\end{table}

\begin{table}[h]
  \centering
  \caption{Results of our framework and without Linear transformer of step segmentation on HHA300.}\footnotesize
  \begin{tabular}{@{}lccccc@{}}
    \toprule
    Method              & Acc$\uparrow$    & Edit$\uparrow$    & F1@\{10,25,50\} $\uparrow$ \\         
    \midrule
    {$Our_{w/o LT}$} & 89.0 & 80.9 & 87.2 86.1 81.9                      \\
    \midrule
    {Our}        & \textbf{89.1} & \textbf{83.3} & \textbf{89.7 89.2 83.0}       \\
    \bottomrule
  \end{tabular}
  \label{tab:linear of hand hygiene}
\end{table}

\begin{table*}[h]
  \centering
  \caption{Results and computational costs of our method with linear transformer and with traditional transformer in step segmentation on HHA300 dataset. $TraT$ denotes the traditional transformer. }\footnotesize
  \begin{tabular}{@{}lcccccc@{}}
    \toprule
    Method              & Acc$\uparrow$    & Edit$\uparrow$    & F1@\{10,25,50\} $\uparrow$  & params($M$)         &FLOPs($G$)   & GPU Mem.   \\
    \midrule
    {Our$_{TraT}$} & \textbf{89.8} & 82.7 & 89.4 89.1 \textbf{83.1} &12.46 &6.14  &$\sim$2.63G                    \\

    {Our}         & 89.1 & \textbf{83.3} & \textbf{89.7 89.2} 83.0 &12.46 &5.77   & $\sim$2.09G      \\
    \bottomrule
  \end{tabular}
  \label{tab:linear and tra}
\end{table*}

\begin{table}[h]
  \centering
  \caption{Comparison of the step based assessment with the traditional method on HHA300 dataset.}\footnotesize
  \begin{tabular}{@{}lcc@{}}
    \toprule
    Method              & Sp. Corr$\uparrow$        & $\mathrm{R}-\ell_{2}(*100)$$\downarrow$\\
    \midrule
    {whole video + MLP}      & 0.390                         & 35.05                       \\
    {segment + MLP}          & \textbf{0.852}     &\textbf{1.07 }                    \\
    \bottomrule
  \end{tabular}

  \label{tab:step with tra}
\end{table}

\begin{table}[h]
  \centering
  \caption{Ablation study of key action scorer of the hand hygiene assessment on HHA300.}\footnotesize
  \begin{tabular}{@{}lcc@{}}
    \toprule
    Method              & Sp. Corr$\uparrow$        & $\mathrm{R}-\ell_{2}(*100)$$\downarrow$\\
    \midrule
    {segment + MLP}          & 0.814     &1.49                     \\
    {segment + KAS(our)}         & \textbf{0.852}     &\textbf{1.07} \\
    \bottomrule
  \end{tabular}

  \label{tab:ass of hand hygiene}
\end{table}

\begin{table}[h]
  \centering
  \caption{Results of our method with the learnable Sigmoid and the Sigmoid in key action scorer (KAS) on HHA300 dataset.}\footnotesize
  \begin{tabular}{@{}lcc@{}}
    \toprule
    Method              & Sp. Corr$\uparrow$        & $\mathrm{R}-\ell_{2}(*100)$$\downarrow$\\
    \midrule
    {KAS(original Sigmoid)}          & 0.821     &1.42                     \\
    {KAS(learnable Sigmoid)}         & \textbf{0.852}     &\textbf{1.07} \\
    \bottomrule
  \end{tabular}

  \label{tab:sigmoid}
\end{table}

\subsection{Ablation Study}

\textbf{\flushleft Effectiveness of the optical flow}.
Because traditional action segmentation methods~\cite{Li,Wang} usually only use RGB features, we introduce optical flow information to improve the discriminative ability of feature representations. Table \ref{tab:optical flow of hand hygiene} shows the comparison results of using only RGB features and RGB features with optical flow features on HHA300 dataset. As shown in the table, the introduction of optical flow information can well assist the task of step segmentation by leveraging motion cues, and thus improve all metrics.

\textbf{\flushleft Effectiveness of the linear transformer}.
We design a multi-stage convolution-transformer network based step segmentation. Table \ref{tab:linear of hand hygiene} shows the comparison results of the multi-stage convolution network and the multi-stage convolution-transformer network on HHA300 dataset. As shown in the table, after embedding the linear transformer, although the improvement in accuracy is very small, other metrics have great improvements, and these metrics are usually used to reflect the ability of the model to reduce over-segmentation. Therefore, the proposed linear transformer based multi-stage model can effectively alleviate the over-segmentation problem. In addition, we also compare the results and computation costs of traditional transformer and linear transformer, as shown in Table \ref{tab:linear and tra}. Base on these results, we can see that the linear transformer based model can achieve similar performance with traditional transformer but has a lower computation cost.

\textbf{\flushleft Effectiveness of the step based assessment}.
To verify the effectiveness of our step based assessment scheme, we compare it with the traditional method that regresses the whole video to a score. To be fair, we use MLP in both methods to regress the score and the results are shown in Table \ref{tab:step with tra}. As can be seen from the table, our step based assessment method is far superior to the traditional method that regresses the whole video to a score. The reason is that traditional method is hard to accurately assess step based tasks, whose qualities are usually determined by key actions in each step. While we propose the step based assessment model which can solve this problem well, thus obtain much better results.

\textbf{\flushleft Effectiveness of the key action scorer}.
To verify the effectiveness of the key action scorer, we evaluate two methods in Table \ref{tab:ass of hand hygiene}. They are the assessment method of the video after segmentation with MLP, and the assessment method of the video after segmentation with key action scorer. The results validate the effectiveness of our design.

\textbf{\flushleft Effectiveness of the learnable Sigmoid}.
We use the learnable Sigmoid instead of original Sigmoid in KAS. To verify the influence of the learnable Sigmoid on KAS, we compare it with original Sigmoid. The experimental results are shown in Table \ref{tab:sigmoid}. The learnable Sigmoid we selected can achieve the best performance, which demonstrates the effectiveness of the learnable Sigmoid in KAS.

\section{Conclusion}
In this work, we present a fine-grained learning framework to perform step segmentation and key action scorer in a joint manner for accurate hand hygiene assessment.  Instead of direct assessment of a whole video, we design a multi-stage convolution-transformer network to generate step segments and then design an action assessment network based on a key action scorer to assess each step. Experimental results show that our framework can accurately assess hand hygiene videos against some state-of-the-art methods. In addition, we contribute a unified hand hygiene video dataset to promote the research and development of hand hygiene assessment. In the future, we will extend our framework to an online version for online hand hygiene assessment to improve the practicability, and expand our dataset to include more challenging scenes.



\bibliographystyle{IEEEtran}
\bibliography{reference}

\vfill

\end{document}